# Leaf Spectral Reflectance Prediction Using Multi-Head Attention Neural Networks

Parastoo Farajpoor[a], Alireza Pourreza[a], Mohammadreza Narimani[a], Ashraf El-kereamy[b], Matthew W. Fidelibus[c]
[a] Digital Agriculture Laboratory, Department of Biological and Agricultural Engineering, University of California, Davis, CA, USA
[b] Department of Botany and Plant Sciences, University of California, Riverside, CA, USA
[c] Department of Viticulture and Enology, University of California, Davis, CA, USA

## ABSTRACT

Accurate modeling of leaf spectral reflectance from physiological and biochemical traits is essential for advancing remote sensing applications in plant science and precision agriculture. Widely used radiative transfer models, such as PROSPECT-PRO, rely on generalized trait–reflectance relationships developed from a wide range of species, which may not fully capture the spectral behavior of specific crops like grapevines. In this study, we developed a trait-to-spectra prediction model using a multi-head attention neural network trained on a grapevine-specific dataset that includes 16 leaf traits measured across multiple varieties, growth stages, and years. The model was evaluated using stratified 5-fold cross-validation and achieved an average coefficient of determination ($R^2$) of 0.84 and normalized root mean squared error (NRMSE) of 1.52%, demonstrating high accuracy and generalizability. When compared to PROSPECT-PRO in forward mode, the neural network exhibited lower mean absolute error (MAE), especially in the near-infrared (NIR) and shortwave-infrared (SWIR) regions. These results emphasize the importance of species-specific modeling approaches and show that integrating biochemical and structural traits into data-driven architectures can significantly improve spectral prediction. The proposed model provides a robust framework for generating accurate leaf-level reflectance data, with potential applications in canopy trait retrieval, vineyard monitoring, and remote sensing-driven crop management.

## 1. INTRODUCTION

Understanding and modeling the relationship between leaf traits and spectral reflectance is fundamental to advancing remote sensing applications in agriculture and plant physiology. Previous studies have shown that various leaf traits strongly influence spectral reflectance [1, 2, 3, 4]. Leaf reflectance plays a key role in radiative transfer models (RTMs), which use this information to estimate canopy-level variables such as biomass, nutrient content, and photosynthetic activity [5, 6, 7, 8]. Because RTMs typically require accurate leaf-level spectral data as input, improving our ability to predict these spectra directly from biochemical and nutritional traits can significantly enhance trait retrieval across spatial and temporal scales.

Among the available leaf RTMs, PROSPECT-PRO [9] has become widely used for estimating leaf spectra based on traits such as chlorophyll (Chl), protein (Prot), and carbon-based constituents (CBC). While it offers a strong foundation for generalized applications, its performance may be limited when applied to specific crops like grapevines. This is mainly because PROSPECT-PRO was developed using a diverse set of plant species, which may not fully capture the unique structural and chemical characteristics of grape leaves.

To address this limitation, we developed a machine learning model specifically trained on grapevine leaf samples collected across different varieties and growth stages. Unlike general RTMs, this model is tailored to the unique reflectance behavior of grapevines and leverages a broader set of traits, including 16 biochemical and nutritional parameters. These traits include nitrogen (N), phosphorus (P), potassium (K), calcium (Ca), magnesium (Mg), manganese (Mn), zinc (Zn), iron (Fe), copper (Cu), boron (B), water content (EWT), leaf mass per area (LMA), anthocyanins (Ant), carotenoids (Car), structural parameter (Nstruct) and Chl, which are known to influence the leaf spectral reflectance across the visible, near-infrared, and shortwave-infrared regions of the spectrum. By directly modeling the spectral response from these traits, our approach takes advantage of available biochemical and nutritional information to produce more accurate spectral predictions, ultimately supporting more precise canopy-level trait estimation through RTMs.

# 2. METHODOLOGY

## 2.1. Data collection

Leaf samples were collected from three grapevine varieties—Sunpreme (white raisin), Flame Seedless (red table grape, pot-grown), and Solbrio (black table grape)— in three consecutive years, from 2021 to 2023, at the University of California's Kearney Agricultural Research and Extension Center (KREC) in Parlier, California. The study encompassed 2,305 leaves (992 unique samples) collected at varying canopy heights and phenological stages (prebloom, bloom, berry set, veraison, harvest) to capture physiological diversity.

Reflectance spectra (350–2500 nm) were acquired using an SVC HR-1024i spectrometer with a Leaf Clip and internal halogen light source. Calibration with a Spectralon disk was performed every five samples, and dark current correction was automated. The spectral data were then standardized to 1 nm resolution (400–2500 nm) via linear interpolation, and a Savitzky–Golay filter (75 nm window, third-order polynomial) was applied to the spectral data to minimize the noise. The resulting spectral reflectance is illustrated in Figure 1.

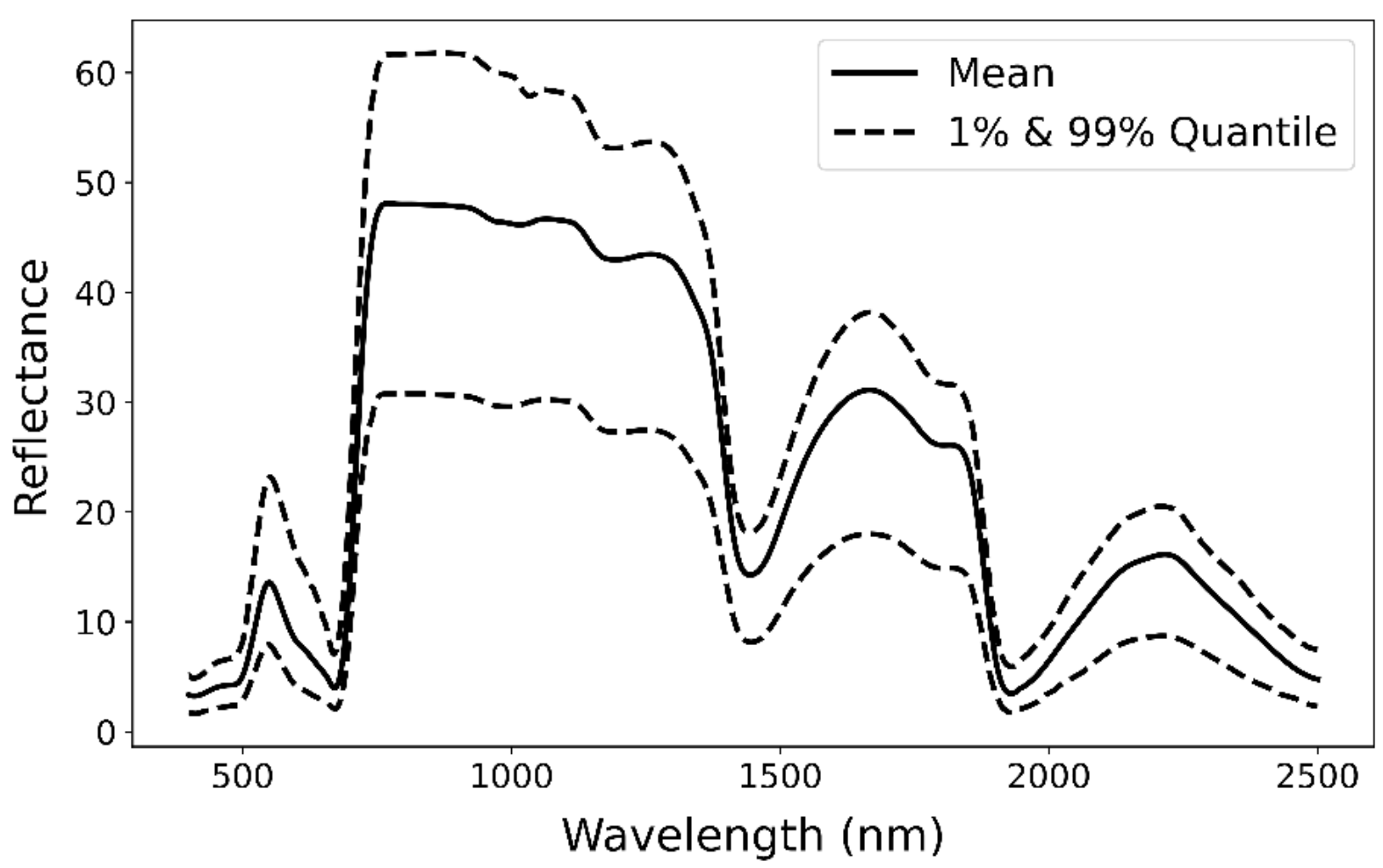

Figure 1. Mean and 1%-99% quantiles of spectral reflectance across all datasets

Post-spectral analysis, weight, and area of leaves were recorded. Chlorophyll (Chl) extraction involved methanol-based processing of 9 mm leaf discs (~0.1 g), stored at –20 °C. The remaining tissue was oven-dried (60 °C) to determine dry mass, enabling calculations of leaf mass per area (LMA) and equivalent water thickness (EWT) using Equations 1 and 2.

$$LMA = \frac{Dry\ weight}{Leaf\ area} \quad (1)$$

$$EWT = \frac{Fresh\ weight - Dry\ weight}{Leaf\ area} \quad (2)$$

Nutrient analysis (N, P, K, Ca, Mg, Zn, Mn, Fe, Cu, B) was conducted by a commercial lab, with data normalized to leaf area for canopy-level scaling applications [10, 11]. The specific leaf traits measured for each grapevine variety and collection year are summarized in Table 1.

Table 1. Summary of 13 measured grapevine leaf traits (#n = Number of samples)

| Year | Variety | #n | Traits | | | | | | | | | | | | |
|---|---|---|---|---|---|---|---|---|---|---|---|---|---|---|---|
| | | | N | P | K | Ca | Mg | Zn | Mn | Fe | Cu | B | Chl | EWT | LMA |
| 2021 | Flame Seedless | 470 | ✓ | | | | | | | | | | | ✓ | ✓ |
| 2022 | Solbrio | 195 | ✓ | | | | | | | | | | | ✓ | ✓ |
| 2023 | Solbrio – Sunpreme | 327 | ✓ | ✓ | ✓ | ✓ | ✓ | ✓ | ✓ | ✓ | ✓ | ✓ | ✓ | ✓ | ✓ |

## 2.2. Imputation of traits

To address missing trait values in the dataset, the PROSPECT-PRO radiative transfer model was used in the inverse mode to estimate missing Nstruct, Ant, and Car in all samples, and Chl in 665 samples. To ensure consistency between measured and estimated Chl values, we assessed the accuracy of PROSPECT-PRO predictions using a subset of 327 samples with direct Chl measurements. A systematic bias was observed, prompting the application of a correction factor. This factor was derived from the inverse of the slope of a best-fit line comparing predicted and measured Chl values (Figure 2a), resulting in a scaling coefficient of 0.75. The adjusted factor was then applied to all PROSPECT-PRO-derived Chl estimates in the remaining samples to improve alignment with observed data.

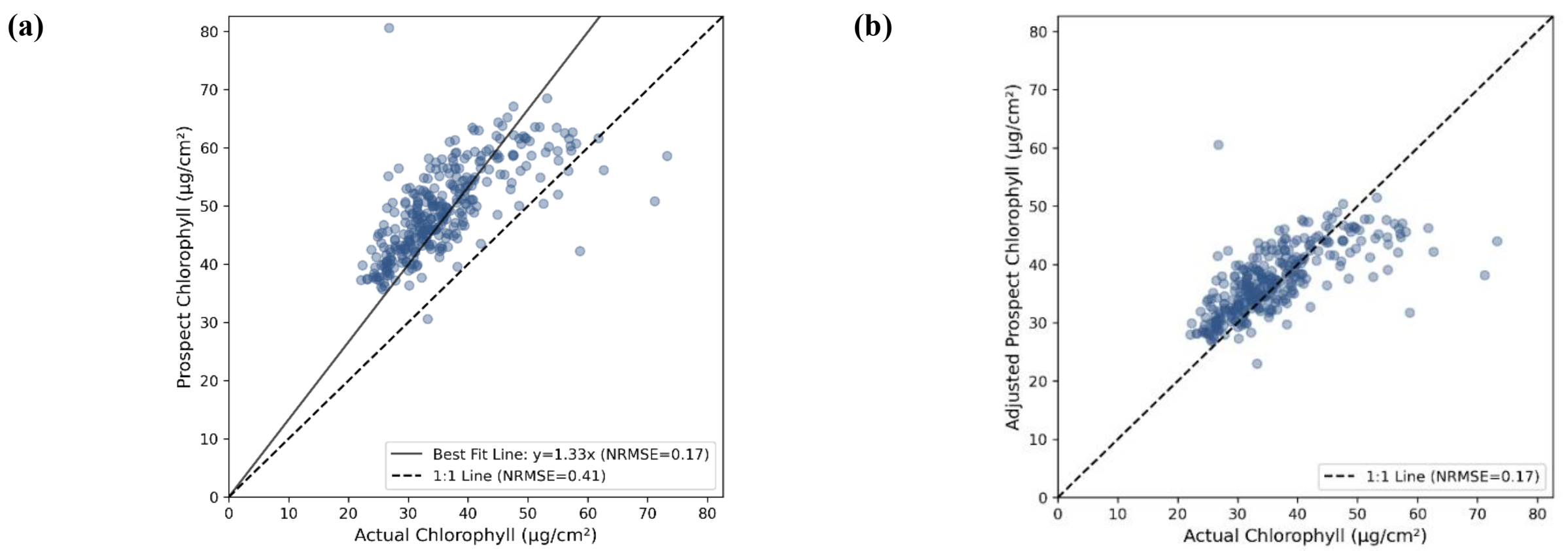

Figure 2. Comparing PROSPECT-PRO estimates and actual Chl values (a) before adjustment and (b) after adjustment

To enable full use of the dataset, missing trait values in 665 samples from the 2021 and 2022 collections were estimated using a convolutional neural network (CNN). First, Principal Component Analysis (PCA) was applied to the hyperspectral data from all the 992 samples. The first 23 principal components, accounting for over 99.99% of the variance, were selected as model inputs to reduce dimensionality while preserving critical spectral information.

The CNN model was trained using the 2023 dataset, with 23 PCs as inputs and all 16 leaf traits as outputs. The model architecture included a 1D convolutional layer (32 filters, kernel size = 3), followed by batch normalization and dropout. This was connected to two fully connected layers (128 and 64 neurons) with ReLU activations and L1/L2 regularization. A final linear layer produced predictions for all traits.

To improve performance, trait values were log-transformed and standardized before training. Predictions were inverse-transformed and exponentiated to return them to their original scale. The model was optimized using Adam (learning rate = 0.001) and trained with mean squared error (MSE) loss calculated from Equation 3:

$$MSE = \frac{1}{n} \sum_{i=1}^{n} (y_i - \hat{y}_i)^2 \quad (3)$$

Where:

- $y_i$ = actual values
- $\hat{y}_i$ = predicted values
- $n$ = number of data points

Five-fold cross-validation was used to evaluate performance, ensuring balanced representation across datasets. Metrics included $R^2$ and NRMSE, calculated as:

$$R^2 = 1 - \frac{\sum_{i=1}^{n} (y_i - \hat{y}_i)^2}{\sum_{i=1}^{n} (y_i - \bar{y}_i)^2} \quad (4)$$

$$NRMSE = \frac{\sqrt{MSE}}{\max(y_i) - \min(y_i)} \quad (5)$$

Where:

- $y_i$ = actual values
- $\hat{y}_i$ = predicted values
- $\bar{y}_i$ = mean of the actual values
- $\max(y_i)$ = maximum of the actual values
- $\min(y_i)$ = minimum of the actual values
- $n$ = number of data points

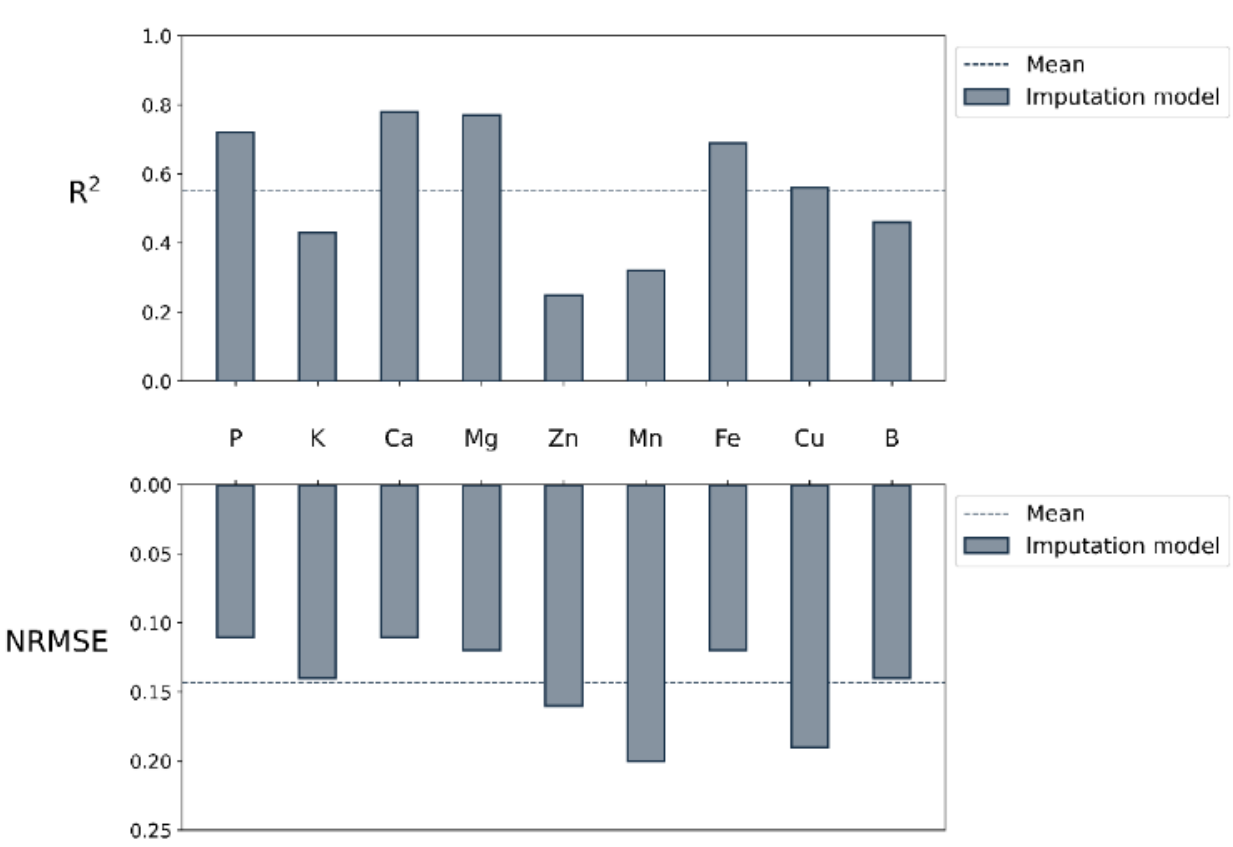

Figure 3. Imputation model performance for various leaf traits

The evaluation metrics for each trait that needed imputation are illustrated in Figure 3. After evaluation, the model trained on the 2023 dataset was used to impute missing traits in the 2021–2022 data.

### 2.3. Model development

We developed a deep learning model incorporating a multi-head attention mechanism to predict leaf spectral reflectance from 16 physiological and biochemical traits. The model was designed to learn complex, nonlinear relationships between the input leaf traits and spectral reflectance outputs spanning 400 to 2500 nm at 1 nm resolution (2,101 bands). A visual summary of the model architecture is presented in Figure 4, illustrating the sequential integration of attention and convolutional layers used to map trait inputs to spectral outputs.

The model architecture began with a fully connected layer (64 units, ReLU activation) to project the 16-dimensional input into a higher-dimensional space. This output was passed through a multi-head self-attention layer with 8 heads and a key dimension of 64, allowing the model to learn interdependencies among the input traits. The encoded representation was then fed into a second attention block and normalized via layer normalization to form a contextualized feature set. These encoded features were further processed through a dense layer, reshaped, and passed through two 1D convolutional layers (64 and 32 filters, kernel size = 20) to model local spectral patterns. The resulting feature maps were flattened and passed through a dense block consisting of a 128-unit fully connected layer, followed by batch normalization and dropout (rate = 0.2) to reduce overfitting. The final output layer was a dense layer with 2,101 units and linear activation, corresponding to the predicted spectral reflectance values for each band.

The model was compiled with the Adam optimizer (learning rate = 0.001) and trained using MSE as the loss function. Training was conducted over 300 epochs with early stopping and learning rate reduction callbacks based on validation loss. Performance was evaluated using 5-fold stratified cross-validation, ensuring proportional representation of all datasets in each fold. Model accuracy was assessed using $R^2$ and NRMSE, calculated between the predicted and measured reflectance spectra.

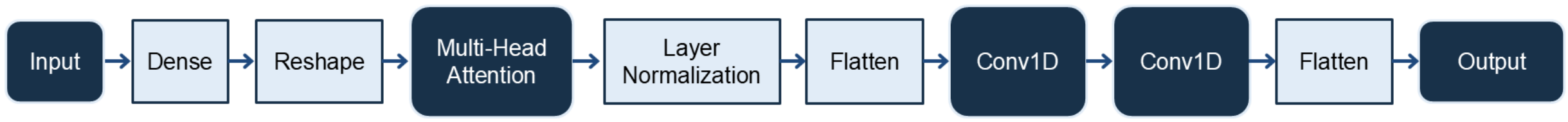

Figure 4. Structure of the spectral prediction model

Following model training, MAE and standard deviation were calculated across all spectral bands to evaluate prediction accuracy and consistency. To further evaluate model performance, predicted spectra were compared with those generated by the PROSPECT-PRO model in the forward mode by calculating the MAE between PROSPECT-PRO predictions and the measured spectra.

## 3. RESULTS AND DISCUSSION

The proposed multi-head attention model demonstrated strong performance in predicting leaf spectral reflectance from 16 physiological and biochemical traits across diverse grapevine samples. In a stratified 5-fold cross-validation setup, the model achieved an average $R^2$ of 0.84 and an average NRMSE of 1.52%, indicating high accuracy and consistency across different folds (Table 2). These results suggest that the model was able to generalize well across different datasets and sampling years, capturing key spectral patterns driven by underlying leaf properties.

Table 2. Performance metrics of the prediction model across all folds of cross-validation

| Fold | $R^2$ | NRMSE (%) |
|---|---|---|
| 1 | 0.81 | 1.61 |
| 2 | 0.84 | 1.72 |
| 3 | 0.79 | 2.02 |
| 4 | 0.85 | 1.12 |
| 5 | 0.89 | 1.12 |
| Average over all folds | 0.84 | 1.52 |

To further evaluate predictive performance, we compared the model's predicted spectra to those generated by the PROSPECT-PRO model operating in forward mode. As shown in Figure 5, the neural network consistently produced lower MAE across most wavelengths. This improvement was especially prominent in wavelength regions sensitive to internal leaf structure and water content, such as the NIR (~700–1300 nm) and SWIR (~1500–2400 nm). The most significant differences were observed in these regions, where PROSPECT-PRO often over- or underestimated reflectance for grapevine leaves. The lower error and narrower standard deviation range of the attention-based model indicate that tailoring the model specifically to grapevine leaf traits improves spectral reconstruction accuracy. In contrast, PROSPECT-PRO, which was trained on a broad range of species, struggled to generalize to the unique structural and chemical properties of grape leaves.

Additionally, the attention mechanism in the proposed model likely contributed to its improved performance by enabling the model to weigh the relative importance of different traits for specific spectral regions [12, 13]. This flexible learning structure allows the model to adapt to complex, nonlinear interactions between traits and reflectance patterns, capturing detailed variations that may vary across wavelengths and trait combinations.

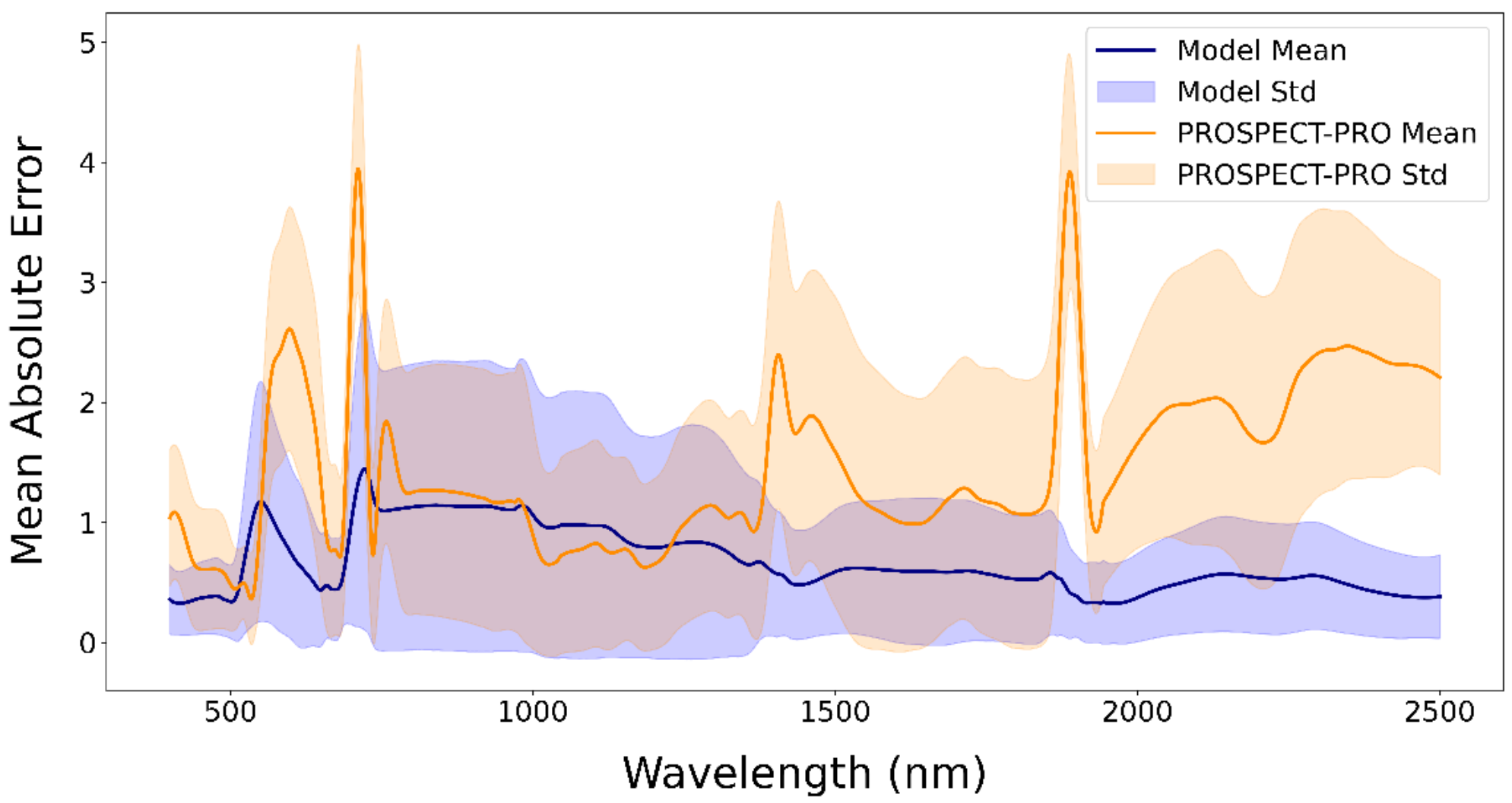

Figure 5. Mean absolute error across wavelengths for model predictions and PROSPECT-PRO estimates

These findings highlight the potential of trait-driven spectral modeling for grapevines and reinforce the value of building crop-specific models. The approach offers a promising alternative for generating accurate spectral data when direct measurement is not feasible and provides a strong foundation for future canopy-level trait estimation using radiative transfer models. By integrating such trait-to-spectra models with radiative transfer frameworks, researchers can simulate realistic canopy-level reflectance and improve remote sensing applications for vineyard monitoring and precision agriculture.

## 4. CONCLUSION

This study presents a grapevine-specific deep learning approach for predicting leaf spectral reflectance directly from physiological and biochemical traits. By training a multi-head attention neural network on a diverse dataset of grapevine leaves collected across varieties and growth stages, the model achieved high predictive accuracy across the 400–2500 nm spectral range. Compared to PROSPECT-PRO, the proposed approach demonstrated lower prediction error, particularly in the near- and shortwave-infrared regions where structural and water-related traits have a strong spectral influence. These findings highlight the advantages of incorporating detailed, crop-specific trait information into spectral modeling and support the development of data-driven tools tailored to specific plant species. The resulting trait-to-spectra model offers a valuable foundation for improving canopy-level trait estimation in viticulture and advancing the integration of machine learning into radiative transfer frameworks for remote sensing applications.

## ACKNOWLEDGEMENT

This work was supported by USDA-NIFA Specialty Crop Research Initiative Award No. 2020-51181-32159 and the California Table Grape Commission. Additionally, the authors would like to thank Momtanu Chakraborty, and Sirapoom Peanusaha from UC Davis Digital Agriculture Lab and Jaclyn Stogbauer and Larry Williams from the Department of Viticulture and Enology, UC Davis, for their help, advice, and directing related papers and materials.

## REFERENCES

[1] Burnett, A. C., Anderson, J., Davidson, K. J., Ely, K. S., Lamour, J., Li, Q., Morrison, B. D., Yang, D., Rogers, A., & Serbin, S. P. (2021). A best-practice guide to predicting plant traits from leaf-level hyperspectral data using partial least squares regression. *Journal of Experimental Botany, 72*(18), 6175–6189. https://doi.org/10.1093/jxb/erab295

[2] Helsen, K., Bassi, L., Feilhauer, H., Kattenborn, T., Matsushima, H., Van Cleemput, E., Somers, B., & Honnay, O. (2021). Evaluating different methods for retrieving intraspecific leaf trait variation from hyperspectral leaf reflectance. Ecological Indicators, 130, 108111. https://doi.org/10.1016/j.ecolind.2021.108111

[3] Ely, K. S., Burnett, A. C., Lieberman-Cribbin, W., Serbin, S. P., & Rogers, A. (2019). Spectroscopy can predict key leaf traits associated with source–sink balance and carbon–nitrogen status. *Journal of Experimental Botany, 70*(6), 1789–1799. https://doi.org/10.1093/jxb/erz061

[4] Yendrek, C. R., Tomaz, T., Montes, C. M., Cao, Y., Morse, A. M., Brown, P. J., McIntyre, L. M., Leakey, A. D. B., & Ainsworth, E. A. (2017). High-throughput phenotyping of maize leaf physiological and biochemical traits using hyperspectral reflectance. Plant Physiology, 173(1), 614–626. https://doi.org/10.1104/pp.16.01447

[5] Asner, G. P., & Martin, R. E. (2008). Spectral and chemical analysis of tropical forests: Scaling from leaf to canopy levels. *Remote Sensing of Environment, 112*(10), 3958–3970. https://doi.org/10.1016/j.rse.2008.07.003

[6] Strow, L. L., Hannon, S. E., De Souza-Machado, S., Motteler, H. E., & Tobin, D. (2003). An overview of the AIRS radiative transfer model. IEEE Transactions on Geoscience and Remote Sensing, 41(2), 303–313. https://doi.org/10.1109/TGRS.2002.808244

[7] Gastellu-Etchegorry, J.-P., Lauret, N., Yin, T., Landier, L., Kallel, A., Malenovský, Z., Al Bitar, A., Aval, J., Benhmida, S., Qi, J., Medjdoub, G., Guilleux, J., Chavanon, E., Cook, B., Morton, D., Chrysoulakis, N., & Mitraka, Z. (2017). DART: Recent advances in remote sensing data modeling with atmosphere, polarization, and chlorophyll fluorescence. IEEE Journal of Selected Topics in Applied Earth Observations and Remote Sensing, 10(6), 2640–2649. https://doi.org/10.1109/JSTARS.2017.2685528

[8] Lei, T., Graefe, J., Mayanja, I. K., Earles, M., & Bailey, B. N. (2024). Simulation of automatically annotated visible and multi-/hyperspectral images using the Helios 3D plant and radiative transfer modeling framework. Plant Phenomics, 6, Article 0189. https://doi.org/10.34133/plantphenomics.0189

[9] Féret, J.-B., Berger, K., de Boissieu, F., & Malenovský, Z. (2021). PROSPECT-PRO for estimating content of nitrogen-containing leaf proteins and other carbon-based constituents. Remote Sensing of Environment, 252, 112173. https://doi.org/10.1016/j.rse.2020.112173

[10] Hill, J., Buddenbaum, H., & Townsend, P. A. (2019). Imaging Spectroscopy of Forest Ecosystems: Perspectives for the Use of Space-borne Hyperspectral Earth Observation Systems. In Surveys in Geophysics (Vol. 40, Issue 3, pp. 553–588). Springer Science and Business Media B.V. https://doi.org/10.1007/s10712-019-09514-2

[11] Zhao, Y., Sun, Y., Lu, X., Zhao, X., Yang, L., Sun, Z., & Bai, Y. (2021). Hyperspectral retrieval of leaf physiological traits and their links to ecosystem productivity in grassland monocultures. Ecological Indicators, 122. https://doi.org/10.1016/j.ecolind.2020.107267

[12] Liu, L., Liu, J., & Han, J. (2021). Multi-head or single-head? An empirical comparison for Transformer training [Preprint]. arXiv. https://arxiv.org/abs/2106.09650

[13] Vaswani, A., Shazeer, N., Parmar, N., Uszkoreit, J., Jones, L., Gomez, A. N., Kaiser, Ł., & Polosukhin, I. (2017). Attention is all you need. Advances in Neural Information Processing Systems, 30. https://doi.org/10.48550/arXiv.1706.03762